%% file: 00_main.tex
\documentclass[letterpaper, journal]{IEEEtran}
\usepackage{times}
\usepackage{epsfig}
\usepackage{graphicx}
\usepackage{amsmath}
\usepackage{amssymb}
\usepackage{comment}
\usepackage{textcomp}
\usepackage[font=footnotesize]{subfig}
\usepackage{multirow}
\usepackage{algorithm}
\usepackage{fixltx2e}
\usepackage[noend]{algpseudocode}
\usepackage{rotating}
\usepackage{cite}

\DeclareMathOperator*{\argmin}{arg\,min}

\usepackage{hyperref}

\usepackage{scalerel}
\usepackage{tikz}
\usetikzlibrary{svg.path}

\definecolor{orcidlogocol}{HTML}{A6CE39}
\tikzset{
  orcidlogo/.pic={
    \fill[orcidlogocol] svg{M256,128c0,70.7-57.3,128-128,128C57.3,256,0,198.7,0,128C0,57.3,57.3,0,128,0C198.7,0,256,57.3,256,128z};
    \fill[white] svg{M86.3,186.2H70.9V79.1h15.4v48.4V186.2z}
                 svg{M108.9,79.1h41.6c39.6,0,57,28.3,57,53.6c0,27.5-21.5,53.6-56.8,53.6h-41.8V79.1z M124.3,172.4h24.5c34.9,0,42.9-26.5,42.9-39.7c0-21.5-13.7-39.7-43.7-39.7h-23.7V172.4z}
                 svg{M88.7,56.8c0,5.5-4.5,10.1-10.1,10.1c-5.6,0-10.1-4.6-10.1-10.1c0-5.6,4.5-10.1,10.1-10.1C84.2,46.7,88.7,51.3,88.7,56.8z};
  }
}

\newcommand\orcidicon[1]{\href{https://orcid.org/#1}{\mbox{\scalerel*{
\begin{tikzpicture}[yscale=-1,transform shape]
\pic{orcidlogo};
\end{tikzpicture}
}{|}}}}

\newcommand{\x}{$\times$}
\begin{document}

\title{A Hybrid Neuromorphic Object Tracking and Classification Framework for Real-time Systems}

\author{Andr\'es~Ussa~\orcidicon{0000-0001-8112-6681}, Chockalingam~Senthil~Rajen~\orcidicon{0000-0002-3132-1830}, Deepak~Singla~\orcidicon{0000-0001-7699-7079},~\IEEEmembership{Member,~IEEE}, Jyotibdha~Acharya~\orcidicon{0000-0002-7167-6775}, Gideon~Fu~Chuanrong~\orcidicon{0000-0002-1871-3274}, Arindam~Basu~\orcidicon{0000-0003-1035-8770},~\IEEEmembership{Senior Member,~IEEE}
        and~Bharath~Ramesh*~\orcidicon{0000-0001-8230-3803},~\IEEEmembership{Member,~IEEE}
\thanks{B. Ramesh is with the N.1 Institute of health, National University of Singapore, Singapore 117456. E-mail: bharath.ramesh03@u.nus.edu.}
\thanks{A. Ussa is with the N.1 Institute for Health, National University of Singapore. G. Fu, and C. SenthilRajen were with the National University of Singapore. D. Singla, J. Acharya, and A. Basu are with the School of Electrical and Electronic Engineering, Nanyang Technological University, Singapore. }
}

\markboth{IEEE Trans. on Neural Networks and Learning Systems}%
{Ussa \MakeLowercase{\textit{et al.}}: Low-power Neuromorphic Vision for Surveillance}

\maketitle

\input{01_abstract}

\begin{IEEEkeywords}
Event-based vision, object tracking, object classification, neuromorphic vision, FPGA implementation, IBM TrueNorth.
\end{IEEEkeywords}

\textit{Source code:} \url{https://github.com/nusneuromorphic/cEOT}

\IEEEpeerreviewmaketitle

\input{02_introduction}

\input{03_methodology}

\input{04_experiments}

\input{05_conclusion}

\bibliographystyle{IEEEtran}
\bibliography{egbib}

\input{06_biography}

\end{document}

%% file: 01_abstract.tex
\begin{abstract}
Deep learning inference that needs to largely take place on the `edge' is a highly computational and memory intensive workload, making it intractable for low-power, embedded platforms such as mobile nodes and remote security applications. To address this challenge, this paper proposes a real-time, hybrid neuromorphic framework for object tracking and classification using event-based cameras that possess desirable properties such as low-power consumption ($5-14 mW$) and high dynamic range ($120 dB$). Nonetheless, unlike traditional approaches of using event-by-event processing, this work uses a mixed frame and event approach to get energy savings with high performance.
\par
Using a frame-based region proposal method based on the density of foreground events, a hardware-friendly object tracking scheme is implemented using the apparent object velocity while tackling occlusion scenarios. The frame-based object track input is converted back to spikes for TrueNorth classification via the energy-efficient deep network (EEDN) pipeline. Using originally collected datasets, we train the TrueNorth model on the hardware track outputs, instead of using ground truth object locations as commonly done, and demonstrate the ability of our system to handle practical surveillance scenarios. As an optional paradigm, to exploit the low latency and asynchronous nature of neuromorphic vision sensors (NVS), we also propose a continuous-time tracker with C++ implementation where each event is processed individually. Thereby, we extensively compare the proposed methodologies to state-of-the-art event-based and frame-based methods for object tracking and classification, and demonstrate the use case of our neuromorphic approach for real-time and embedded applications without sacrificing performance. Finally, we also showcase the efficacy of the proposed neuromorphic system to a standard RGB camera setup when simultaneously evaluated over several hours of traffic recordings.
\end{abstract}

%% file: 02_introduction.tex
\section{Introduction}
\label{sec:intro}
Real-time object tracking consists of initializing candidate regions for objects in the scene, assigning them unique identifiers and following their transition. It is a common requirement to further perform classification over the tracked object. These capabilities of object tracking and classification are valuable in applications like human-computer interaction \cite{Jacob2003}, traffic control \cite{Hsieh2006}, medical imaging \cite{Meijering2006} or video security and surveillance \cite{Hampapur2005}. Current methodologies for surveillance tasks use standard cameras that acquire images or frames at a fixed rate regardless of scene dynamics. Consequently, background subtraction used to retrieve candidate regions-of-interest for tracking is a computationally intensive step, which is also affected by changes in lighting \cite{Piccardi2004}. On the other hand, deployment of cameras with higher frame rate involves a drastic increase in power requirements \cite{Barnich2011}, besides increased demands in memory and bandwidth transmission. Therefore, the frame-based paradigm tends to be intractable for embedded platforms/remote surveillance applications \cite{Basu2018, Cohen2018, frame1, frame2, frame3}.
\par
As an emerging alternative to standard cameras, event cameras acquire information of a scene in an asynchronous and pixel independent manner, where each of them react and transmit data only when intensity variation is observed. This provides a steady stream of events with a very high temporal resolution (microsecond) at low-power ($5-14mW$), reducing redundancy in the data with improved dynamic range due to the local processing paradigm. In particular, there is no significant need for background modeling, since a static event camera will only generate events corresponding to moving objects, thereby naturally facilitating tracker initialization. All these features are well suited for visual tracking applications but demand the use of algorithms designed to handle asynchronous events.
\par
An event-by-event approach is dominantly seen in the literature for object tracking and detection using neuromorphic vision sensors \cite{Lagorce2014, Glover2017, Ramesh2019, event1, event2}. The aim of these methods is to create an object representation based on a set of incoming events and updating it dynamically when events are triggered. Although these methods can be effective for specific applications, they often require high parametrization \cite{Lagorce2014, Glover2017} or are not effective for tracking multiple objects \cite{Ramesh2019}.
\par
Similar to the above works, \cite{Dardelet2018} is an event-by-event approach for object tracking applications that performs a continuous event-based estimation of velocity using a Bayesian descriptor. Another example is \cite{Ramesh2018}, which proposes event-based tracking and detection for general scenes using a discriminative classification system and a sliding window approach. While these methods work intuitively for objects with different shapes and sizes, and even obtain good tracking results, they have not been implemented under a real-time operation requirement.
\par
 \begin{figure*}[t]
 \begin{center}
    \includegraphics[width=0.95\linewidth]{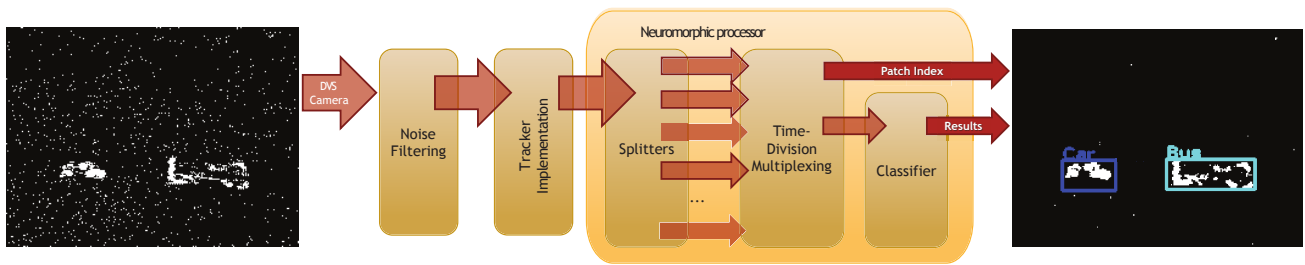}
 \end{center}
   \caption{Block diagram of the real-time neuromorphic surveillance system.}
 \label{fig:teaser}
 \end{figure*} 
\par
In contrast to the above methods, an aggregation of incoming events can be considered at fixed intervals instead of processing events as they arrive. This produces a more obvious representation of the scene (a ``frame''), and allows an easier coupling with traditional feature extraction and classification approaches \cite{Ni2015, Hinz2017, Iacono2018}. In \cite{Hinz2017}, asynchronous event data is captured at different time intervals, such as $10 ms$ and $20 ms$, to obtain relevant motion and salient information. Then, clustering algorithms and Kalman filter are applied for detection and tracking, achieving good performance under limited settings. Other examples of event-based frames along with variations in sampling frequency and recognition techniques are \cite{Ni2015, Iacono2018}, which show the potential of this approach for detection. Taking an important step forward for real-time and embedded applications, we leverage the low-latency and high dynamic range of event cameras interfaced to an FPGA processor for object tracking, followed by object classification on a neuromorphic chip, to provide an end-to-end neuromorphic framework, as shown in Fig.~\ref{fig:teaser}.

In essence, the focus of our approach is to build a real-time and embedded system that takes advantage of a stationary event camera, thereby picking up only moving objects and not being specific to background conditions. To this end, we use a hybrid approach that is different from purely event-based or purely frame-based approaches. First, the asynchronous events are accumulated into a binary image and an overlap-based tracking is performed on these frames. For subsequent object classification, the frames are converted back to spikes for efficient processing on the IBM neuromorphic chip. As shown in the experiments, the hardware-friendly tracker performs significantly better and requires far less resources (7\x less memory and 3\x less computations) than the popular multi-object event-based mean shift (EBMS) tracker \cite{Delbruck2013}. 
Additionally, we compare the performance of the proposed neuromorphic system to a standard RGB camera setup when simultaneously evaluated over several hours of traffic recordings at three different locations. This is of immense importance when using our fully embedded system for remote surveillance applications where long battery life of the sensor node is critical without sacrificing performance. 
\par
This paper is an extended version of the work initially published in BMVC Workshops 2019 \cite{Authorsbmvc19}. Novel contributions over \cite{Authorsbmvc19} include the event-based tracker extension (Sec.~\ref{sec:newone}) and comparison to the  state-of-the-art event and frame-based trackers. We have also evaluated the performance of the proposed tracker on recordings from various commercially available neuromorphic vision sensors (NVS). Additionally, an extensive comparison is made between the TrueNorth classification output to state-of-the-art classification frameworks, including the Spiking Neural Network (SNN) model trained using the method proposed in \cite{shrestha2018slayer}, and also compared with pre-trained models via transfer learning (Section \ref{sec:statecompare}) \cite{DBLP:journals/corr/HeZRS15}.

%% file: 03_methodology.tex
\section{Methodology}
The DAVIS camera events \cite{Brandli2014} are utilized through the formation of frames for the task of tracking vehicles and humans on an urban landscape. Thus, the tracker performance hinges on the ability to capture frames at a rate much faster than the dynamics of the scene, thereby taking advantage of the low-latency of event cameras. The frames obtained are median filtered and region proposals are extracted from two 1-D histograms along $X$ and $Y$ directions for tracking. The tracker uses centroids and Euclidean distances to monitor up to eight objects simultaneously, while classification is performed on these locations using IBM's TrueNorth neuromorphic chip \cite{Merolla2014} to assign one of the following classes: cars, motorbikes, buses, trucks and humans. The filtering and region proposal of the tracker follows the existing method in \cite{Authors} and this work additionally considers occlusion, track velocity calculation and smooth interpolation between two instances of tracking. The full system is embedded on FPGA hardware and interfaced to IBM's TrueNorth chip.
\subsection{Object Tracking}
\label{sec:EOT}
This work proposes a simple, hardware-friendly tracker, termed as events overlap tracker (EOT),  consisting of a series of steps, namely: track assignment, merging and post-processing. The core function resides in the track assignment task, similar to a Kalman filter update, while the merging and post-processing steps deal with the occlusion and track assignment issues. Each track output is defined by a set of properties: (1) The top-left location of the tracked object (\emph{x} and \emph{y} coordinates); (2) The width and height of the tracked object (\emph{w} and \emph{h}); (3) The velocities $v_x$ and $v_y$ of the object; (4) The tracker state (\emph{free}, \emph{tracking}, or \emph{locked}) and (5) A unique ID. The \emph{free} state indicates that the tracker is in stand-by and no region is currently assigned to it. The \emph{tracking} state indicates that the tracker has matched with a region proposal once. The \emph{locked} state indicates that the tracker has matched a region proposal in at least two consecutive frames and is currently ``locked'' on an object. Since only \emph{locked} trackers are classified, having a \emph{tracking} state allows to filter noisy tracks and reduce the burden on the classifier.
\subsubsection{Track Assignment}
The EOT track assignment procedure can be briefly summarized as follows. As a region proposal, defined by its coordinates, $r_j^{new} = \{x_j^{new}, y_j^{new}, w_j^{new}, h_j^{new}\}$, is received as input, its overlap area with respect to all active trackers $T_j^k = \{x_j^k, y_j^k, w_j^k, h_j^k\}$ is measured, where $k=1,\cdots,N$ indicates the track IDs and $j$ the frame instance. If their overlap is higher than the \emph{track assignment ratio} $O_{th}$, the region proposal is then assigned to the respective tracker ID. Otherwise, the region proposal is assigned to a \emph{free} tracker. 
\par
To begin with the track assignment, the tracker's new position is estimated based on its previous velocity, and the resulting region is evaluated against the region proposal. The assignment is evaluated based on the calculation of the \emph{overlap area} $O_{A}$ between two regions, as defined in \eqref{eq:queue1}. 
\begin{multline}
O_{A} = (\max(0, \min(x_j^k + w_j^k, x_j^{new} + w_j^{new}) - \\ \max(x_j^k, x_j^{new})))
  \times (\max(0, \min(y_j^k + h_j^k, y_j^{new} + h_j^{new}) - \\ \max(y_j^k, y_j^{new})))
\label{eq:queue1}
\end{multline}
\subsubsection{Tracker Update}
In our EOT implementation, a tracker assignment is made when the overlapping area is higher than a $O_{th}$ of 20\%. Subsequently, the tracker properties and state are updated. If the current state is \emph{tracking}, then it is updated to \emph{locked}, and if it was already in \emph{locked} state, it will remain as it is. After a successful assignment, each tracker region is updated using a weighted average as stated in \eqref{eq:queue2}, for each of the spatial elements, where $\alpha$ is the weighting degree coefficient.  
\begin{equation}
T_j^k = (1 - \alpha) \cdot r_j^{new} + \alpha \cdot (T_{j-1}^{k} + v_{j-1}^{k} \cdot \Delta t)
\label{eq:queue2}
\end{equation}
where $v_{j-1}^{k}$ refers to the velocity of the track (in pixels/s) at previous frame instance $j-1$, and $\Delta t = t_{j}-t_{j-1}$. The velocity, shown in \eqref{eq:vel1}, is also then averaged analogous to the position update. Similarly, it is also applied for the \emph{y} direction.
\begin{equation}
\label{eq:vel1}
v_{j}^k(x) = (1 - \alpha) \cdot \frac{(x_j^{new} - x_{j-1}^k) + (w_j^{new} - w_{j-1}^{k})}{\Delta t} + \alpha \cdot v_{j-1}^{k}
\end{equation}
\par
During the tracks assignment, in cases where different region proposals are assigned to the same tracker or vice versa, a merging between the pertinent regions is applied. If more than one region proposal, $r_j^{new_1}$ and $r_j^{new_2}$, is assigned to the same tracker $T_{j-1}^k$, non-maximal suppression is performed among the common regions and the tracker region, to group the rectangles and assign it to $T_{j}^k$. On the other hand, if there is a region proposal that is assigned to more than one tracker, an occlusion check is performed among all the valid trackers.
\subsubsection{Occlusion Model}
\label{sec:eot_occ}
Considering that the objects to be tracked display a wide range of sizes and often follow opposite directions or move at different speeds, the case in which an object occludes another occurs regularly. In occlusion scenarios, the event frame would show a bigger region than the individual objects without a clear boundary between them. In other words, the trackers under evaluation will overlap after one or two steps in the future based on the estimated velocity. Before an occlusion occurs, let us denote the implicated trackers as $T_j^{a}$ and $T_j^{b}$, and their size before occlusion as $(w_o^{a},h_o^{a})$ and $(w_o^{b},h_o^{b})$, respectively. Based on the trackers' original sizes, their velocity, direction and the combined area after occlusion, it is possible to approximate their positions during the occluded frames. For this, a set of conditions is determined: trackers' common direction $cd = v_j^{a} + v_j^b > v_j^a \vee v_j^b$, width increase $wi = (w_j^a > w_{j-1}^{a})$ and highest velocity object $hvo = abs(v_j^{a}) > abs(v_j^{b})$.
\par
While the occlusion is occurring, the change in width of the merged tracks, i.e. $wi$, is used as the criteria to determine whether the affected tracks are coming together ($wi=False$) or getting apart ($wi=True$). In particular, for a tracker $T_j^{a}$, it remains as the current region proposal when $wi$ is $False$, $(x_j^{a}, y_j^{a}, w_j^{a}, h_j^{a})\leftarrow (x_{j}^{new}, y_{j}^{new}, w_{j}^{new}, h_{j}^{new})$, or otherwise when $wi$ is $True$, the track is equal to its original size in the region proposal, $(x_j^{a}, y_j^{a}, w_j^{a}, h_j^{a})\leftarrow (x_{j}^{new}+w_{j}^{new}-w_o^{a}, y_{j}^{new}+h_{j}^{new}-h_o^{a}, w_o^{a}, h_o^{a})$. This situation occurs when the objects are moving in opposite directions ($cd=False$), or when they move in a common direction and the velocity of the tracker under evaluation, $T_j^{a}$, is faster than its pair ($hvo=True$). On the other hand, if $hvo$ is $False$, then the track is intuitively set as the other component of the region proposal, $(x_j^{a}, y_j^{a}, w_j^{a}, h_j^{a})\leftarrow (x_{j}^{new}, y_{j}^{new}, w_o^{a}, h_o^{a})$.
\subsubsection{Cleanup}
Finally, the post-processing step removes trackers that no longer match any of the region proposals. This is carried out by comparing the \emph{current} state of the tracker with its \emph{past} state. If a tracker was previously set as \emph{locked} or \emph{tracking} state, and in the current frame it no longer exists, then it is likely that the object is lost. However, since a region proposal can be inconsistent through time, due to the hardware noise from the DAVIS, it is important not to set the tracker free instantly. An intermediate \emph{maximum unlocks} state is used to determine when a tracker is lost for several consecutive frames, and only in that case, it is set \emph{free}. Additionally, an out-of-bounds check is performed to release trackers when objects leave the scene.
\par
Note that although the EOT tracking is discontinuous in time, the location and size of the tracked object can be estimated continuously, allowing the size and location of the object to be determined in-between the frames. In other words, any time $t$ satisfying $t_{1}^{k}<t<t_{n}^{k}$,
\begin{equation}
\begin{array}{l l l}
j &= &\underset{t_{i}^{k}>t}{\argmin_i}(t_{i}^{k})\\
\\
\lambda &= &({t-t_{j-1}^{k}})/({t_{j}^{k}-t_{j-1}^{k}})\\
T &= & T_{j-1}^{k} + \lambda(T_{j}^{k}-T_{j-1}^{k})
\end{array}
\end{equation}
where $j$ is the index of the \textit{closest track} and $\lambda$ is the interpolation factor for time $t$. Using the above equations, the location $(x, y)$ and size $(w,h)$ can be calculated for an interpolated track $T = \{x, y, w, h\}$ at any time $t$. This feature is useful for continuous-time EOT implementations as described next for certain applications. 
\subsection{Continuous-time EOT}
\label{sec:newone}
An events overlap tracker in continuous-time is also presented using the fundamental concepts behind EOT while aiming to fully leverage the low latency nature of event-based cameras. The tracking stage processes each event individually and can be broken into two substages. The first substage assigns each event to one of the trackers (or to no tracker), and the second substage updates the assigned tracker using the new event information and determines whether the tracker status is \textit{active} or \textit{inactive}. Periodically, (typically $25 ms$) a cleanup operation is performed to update old trackers and to merge overlapping trackers. Finally, an occlusion check is performed to improve tracking performance for objects overlapping each other.
\par
We define the $i$th tracker $T_i$ as
\begin{equation}
T_i = \{x_i, y_i, dx_i, dy_i, active_i, isi_i, t_i\}\\
\end{equation}
where $x_i, y_i$ is the current tracker location, $dx_i, dy_i$ is the half-width and half-height of the rectangular tracker, $active_i$ is boolean, indicating whether the tracker is active or not, $isi_i$ is the average inter spike interval, and $t_i$ is the time at which the tracker was last updated.

\subsubsection{Track Assignment}
For each event, the x-direction and y-direction distances to each tracker center are computed and compared to the tracker rectangle size. If the event lies within the tracker rectangle for an active tracker, then it is assigned to that active tracker. More formally, the event $e_i$ will be assigned to the first tracker $T_j$ which is found such that
\begin{equation}\label{eq:inside_tracker}
\begin{array}{l l l}
active_j &=& true\\
|x_j-x_i| &\leq& dx_j\\
|y_j-y_i| &\leq& dy_j\\
\end{array}
\end{equation}

If no such tracker $T_j$ exists, then the event does not lie close enough to any active tracker and will instead be assigned to the nearest inactive tracker. More formally, $T_k$ according to
\begin{equation}
\min_{k |active_k = false}{\sqrt{(x_k-x_i)^2+(y_k-y_i)^2}}\\
\end{equation}

It is possible that an event gets matched to no trackers. This can occur when all trackers are active, but the event is not close enough to any of the active trackers to be allocated to that one of them, in which case the event is omitted.

\subsubsection{Tracker Update}\label{sec:track:update}
The tracker update step consists of updating the assigned tracker's location, and keeping record of the average time between events assigned to the tracker. These quantities are updated using an exponential moving average. If the tracker $T_j$ is being updated by event $e_i$, the position update takes the form
\begin{equation}
\begin{array}{l l l}
x_j &\leftarrow &\alpha x_j + (\alpha-1)x_i\\
y_j &\leftarrow &\alpha y_j + (\alpha-1)y_i\\
\end{array}
\end{equation}
where $\alpha$ is typically set to 0.95.

If the event lies within the tracker region, as determined by satisfying the last two conditions in \eqref{eq:inside_tracker}, then the inter spike interval is updated as
\begin{equation}
isi_j \leftarrow \alpha_tisi_j +(1-\alpha_t)(t_i-t_j)\\
\end{equation}
where $\alpha_t$ is typically set to 0.9.

The time $t_j$ is also modified according to the tracker's last update as $t_j \leftarrow t_i$.

Finally, we perform a check to ascertain whether the tracker should be considered active or not. The check compares the average inter spike interval per pixel within the tracker rectangle to a fixed threshold. If the interval is small enough, the tracker is considered active. The average is calculated as $isi_j\times dx_j\times dy_j$.

If the quantity above is less than the active threshold $\Theta_{active}$, then the tracker is marked as active by setting $active_j\leftarrow true$, otherwise we mark the tracker as inactive using $active_j\leftarrow false$.

\subsubsection{Periodic Cleanup}
A periodic cleanup of trackers is performed every $25 ms$. Without the cleanup, trackers would only be updated when events are assigned to them, which leads to the possibility of an active tracker remaining active indefinitely. The cleanup updates each tracker by generating a false event on the same location as the tracker and at the time of the cleanup, and following the same update process as described in Section~\ref{sec:track:update}.

Once all the trackers are updated, a check is performed to merge overlapping active trackers. Two trackers $T_i$ and $T_j$ are merged only if
\begin{equation}
\begin{array}{l l l l l}
active_i &= &active_j &= &1\\
|x_i-x_j| &\leq &dx_i+dx_j\\
|y_i-y_j| &\leq &dy_i+dy_j.\\
\end{array}
\end{equation}
The new tracker location is the mean of the location of the two trackers being merged. The size of the merged tracker is either the size of the larger tracker, or the sum of the sizes of the two trackers, depending on whether the center of the smaller tracker lies within the rectangular region defined by the larger tracker. Once a tracker is merged into another, it is randomly re-initialized.
\subsubsection{Occlusion Model}
An occlusion handling stage is also implemented in this pipeline taking as reference the occlusion check procedure used in Section \ref{sec:eot_occ}. The first step is to detect an actual occlusion is happening. As there is no concept of a ``region'' in this purely event-based method, possible occlusions are initially detected when an incoming event is being matched to the available trackers (Section ~\ref{sec:track:update}). If an event gets matched to more than one active tracker, then the trackers involved are considered for further checks.

The trackers under occlusion are compared against each other to check for minimum conditions of velocity and covariance. This step allows to filter out false occlusions from being processed. The velocity conditions are shown in \eqref{eq:occ_Da} and \eqref{eq:occ_Db}, and their purpose is to ensure there is a minimum velocity difference between the trackers. The covariance condition is shown in \eqref{eq:cov} and its purpose is to verify that the estimated error of the trackers is low.

\begin{equation}
\label{eq:occ_Da}
D_{\alpha} = \begin{Bmatrix}\begin{vmatrix}Vx_{i} - Vx_{j}\end{vmatrix} > V_{\alpha} \mid\ sign(Vx_{i}) = sign(Vx_{j}) \end{Bmatrix}
\end{equation}
\begin{equation}
\label{eq:occ_Db}
D_{\beta} = \begin{Bmatrix}\begin{vmatrix}Vx_{i} - Vx_{j}\end{vmatrix} > V_{\beta} \mid\ sign(Vx_{i}) \neq sign(Vx_{j}) \end{Bmatrix}
\end{equation}
\begin{equation}
\label{eq:cov}
P_{d}  = diag(P_i) < P_T \wedge diag(P_j) < P_T
\end{equation}
where $V_{\alpha}$ and $V_{\beta}$ are velocity threshold values for the same direction and opposite direction cases, respectively, $P_T$ is the covariance threshold value, and $diag(P_i)$ and $diag(P_j)$ are the diagonal sum of the covariance of each tracker being compared. Then, if $P_d \wedge D_{\alpha} \vee D_{\beta} = 1$, the trackers concerned are considered for the next occlusion detection step.

Following the occlusion detection logic from Section \ref{sec:eot_occ}, the last action is to check for overlapping between the trackers at one or two steps in the future. However, there is no concept of ``timestep'' in this method since there is no frame creation. Therefore, for this purpose, an occlusion timestep $O_t$ was set to perform this verification. The position of a tracker box on subsequent timesteps is calculated as in \eqref{eq:occ_ts}.
\begin{equation}
\label{eq:occ_ts}
\begin{split}
x_i &= x_i + Vx_i \cdot O_t \cdot n\\
y_i &= y_i + Vy_i \cdot O_t \cdot n
\end{split}
\end{equation}
where $n$ is the number of timesteps to take into consideration. Then, if an overlap is found at either $n=1$ or $n=2$, the trackers are flagged as occluding.

While in occlusion state, the processing of occluding trackers changes. First, while in normal state  one incoming event only updates a single tracker, in occlusion state both implicated trackers are updated based on a single matching event. Second, the box size of the involved trackers is preserved throughout the occlusion. This keeps the boxes from over expanding due to the increase of events in the proximity result of the overlapping. Lastly, the velocity of the trackers before occlusion is kept fixed during the occlusion period. This is used to estimate the motion of the tracks during occlusion and to attempt to recover the locked objects after separation.

It is worth mentioning that the occlusion detection process explained above is applied even on trackers marked as occluding. Then, when evaluating \eqref{eq:occ_ts}, if the occluding trackers no longer show to be overlapping at future timesteps, their flag is then removed and the occlusion is assumed concluded.

\subsubsection{Implementation}
A C++ implementation of the tracker was written and evaluated. It is capable of performing tracking far faster than real-time, allowing us to generate tracking information for the entire dataset in approximately 15 mins. The C++ tracker output was written to a file and was used to generate the results shown later.

\subsection{Object Classification}
This section describes the process of object classification on the TrueNorth (TN) chip using IBM's Energy Efficient Deep Network (EEDN) pipeline. For classifying multiple object tracks, the EEDN pipeline is time-multiplexed to handle eight different objects pseudo-simultaneously. This approach is useful when there are only a few objects to classify in a single frame, as neurons used for spike generation will scale linearly with the number of objects. However, in general, the input to the TrueNorth chip can either be raw spikes data or images that are subsequently converted to spikes on the host FPGA.
\par
Feeding the raw spikes to the TrueNorth is not a good choice because it will degrade the system performance and increase latency when TN is placed at a gateway node. In such a case, the workload for TN will be very high since it receives tracks for classification from multiple sensor nodes. As opposed to a continuous streaming of all events, it would be more efficient in terms of time and power to periodically transmit a binary image to TrueNorth. For instance, an uncompressed 32$\times$32 pixel binary image would require a memory of 32$\times$32 bits. In contrast, a continuous spike stream would require a memory of approximately 24 bits per spike. Thus, assuming that each image contains more than $32\times32/24=43$ spikes, the binary image transmission method is more efficient. Note that each TrueNorth classification image contains only the tracked regions of the full binary image, which leads to the following integration issues to be considered.
\par
The binary images from each object track are of different sizes, but TrueNorth EEDN requires fixed size images for the CNN based classification. Thus, we resize each binary track image to a fixed size, $42\times42$, before streaming it to TrueNorth. Additionally, EEDN expects a typical RGB image as input, where the first layer of convolution is performed on the host FPGA using multi-bit inputs and the resulting image is thresholded to generate spikes to be fed to the TrueNorth chip. Since we are using 1-bit binary images, any ones in the image can simply be treated as spikes and can then be passed through all convolution layers on TrueNorth, bypassing the need to generate spikes using the host FPGA. This approach is in line with the intended use of IBM's EEDN tools and allows us to better leverage their pipeline. The CNN model ran on TrueNorth is 15 layers deep, similar to the network used in \cite{Esser2015}.
\par
The first convolution layer is performed on the host FPGA, which is then thresholded to generate spike ticks that are subsequently fed to the TrueNorth chip. We trained the network with a learning rate of 20 for first 500 epochs, followed by learning rate of 2 for the next 500 epochs and finally a learning rate of 0.2 for the next 500 epochs. The actual core count used for the trained CNN network was 2681 out of the available 4096 cores. 

 \begin{figure*}[t]
 \begin{center}
    \includegraphics[width=0.75\linewidth]{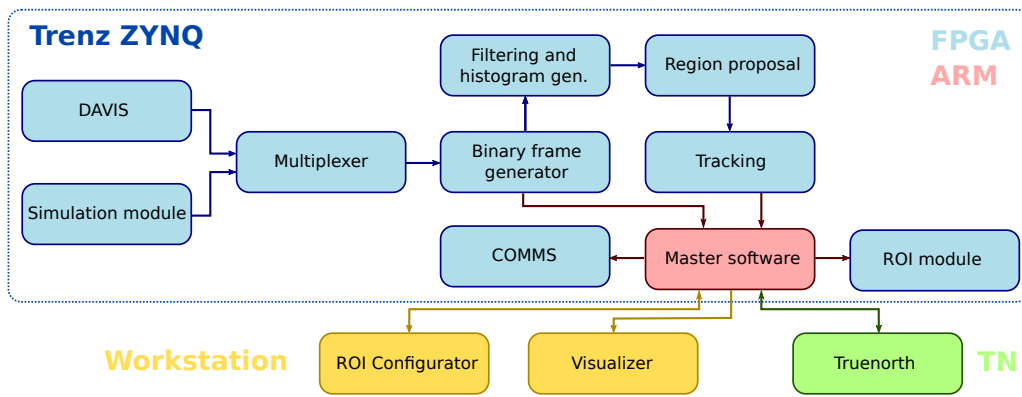}
 \end{center}
   \caption{System flow diagram of the end-to-end neuromorphic surveillance system consisting of the DAVIS vision sensor, an FPGA and ARM processor on the Trenz carrier board that is directly interfaced to IBM TrueNorth.}
 \label{fig:flowdiag}
 \end{figure*}

\subsection{Hardware implementation}
The hardware implementation contains a DAVIS240C event-based sensor, a Trenz TE0720 processing platform that includes an FPGA and an ARM processor, IBM's TrueNorth neuromorphic chip with 4096 cores, and a workstation solely for visualization. The overall operation of the system consists of acquisition of events from the camera, the processing of these events to extract tracked objects, the transmission of these tracked regions to the neuromorphic chip for classification, and the object detection visualization, as shown in Fig.~\ref{fig:flowdiag}.  
\par
The Trenz TE0720 module includes an ARM dual-core Cortex-A9 processing system that runs Ubuntu Linux 16.04 LTS and it handles the interface between the FPGA, the TrueNorth and the workstation for visualization. The communication (COMMS) module implemented on the FPGA allows retrieving the EOT information and sending it to the visualizer after streaming it to TrueNorth for obtaining the classification result. The simulation interface allows sending events from binary files to verify the behavior of the developed modules with prerecorded data. 
\subsection{Power Consumption}
The DAVIS can operate at a few milliwatts ($10 mW$), the Trenz operates at about $422mW$ excluding the system's base power for running Ubuntu operating system, and the TrueNorth chip operates at 100mW. Overall, the power consumption of our system is about $550mW$, which is $3\times$ lower than performing inference on the edge for a similar deep learning network. In particular, an Inception-v3 network running on Google's edge TPU in operation consumes about $1.7W$ \cite{NANOCHIPS2030, GoogleEdgeTPU}. On the other hand, Brix embedded systems (Intel-i7 processor with 8G RAM), like the one described in \cite{Smedt2015} for real-time object tracking using frame-based sensors, consume about 22W, which is $40\times$ more than our implementation. Note that the Trenz Zynq module is a powerful and flexible development tool, but far exceeds the utilities compared to SmartFusion FPGAs that allow sleep modes, non-volatile configuration memory, and have much lower overall power consumption. In other words, there is significant room for very low power ($<50mW$) implementation of our framework with appropriate hardware choices and development efforts. 

%% file: 04_experiments.tex
\section{Experiments}
\begin{figure}[t]
\centering
\subfloat[]{{\includegraphics[width=7cm]{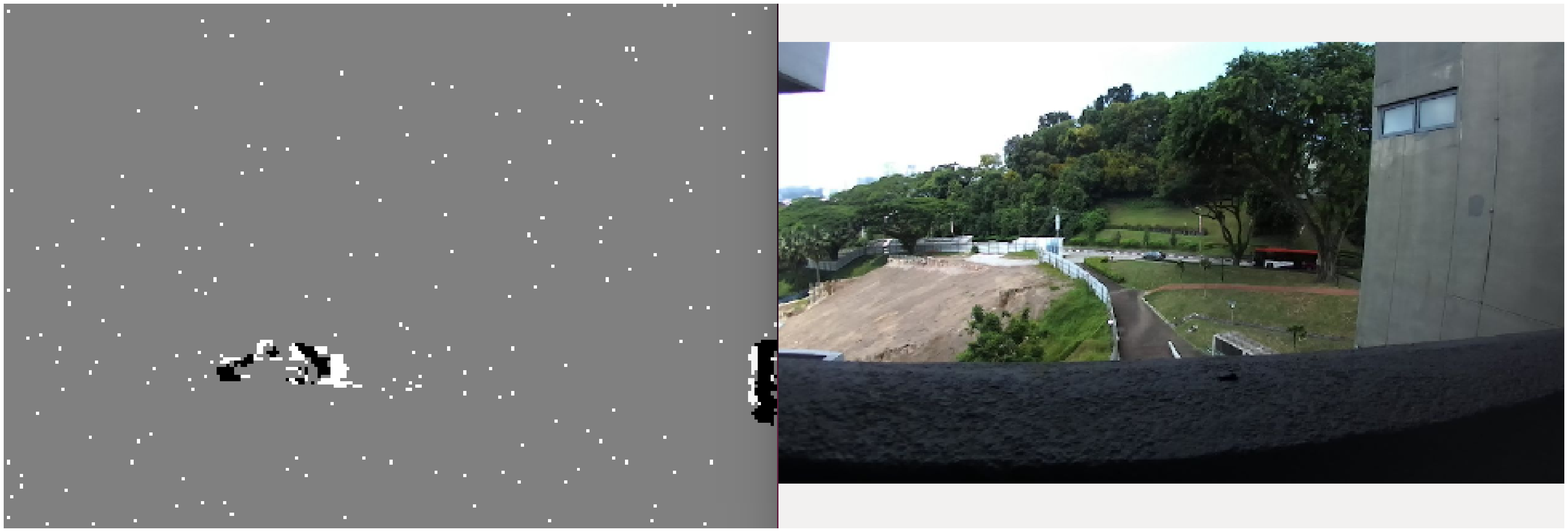}}}
\vspace{0.1em}
\subfloat[]{{\includegraphics[width=7cm]{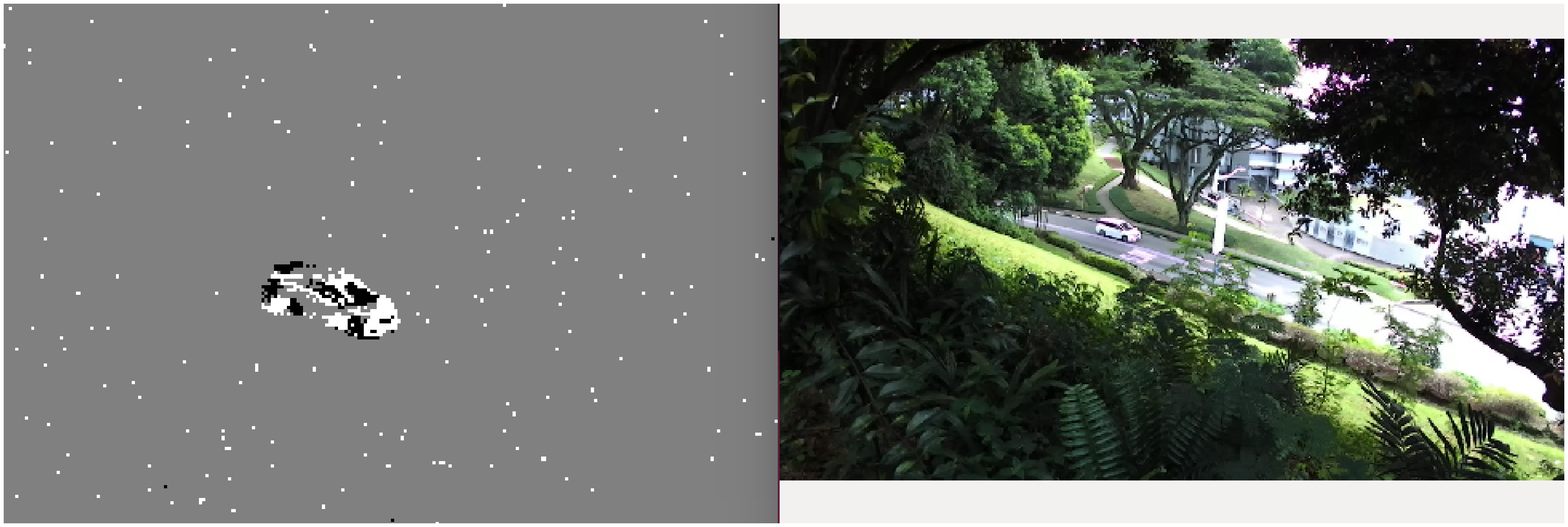}}}
\vspace{0.1em}
\subfloat[]{{\includegraphics[width=7cm]{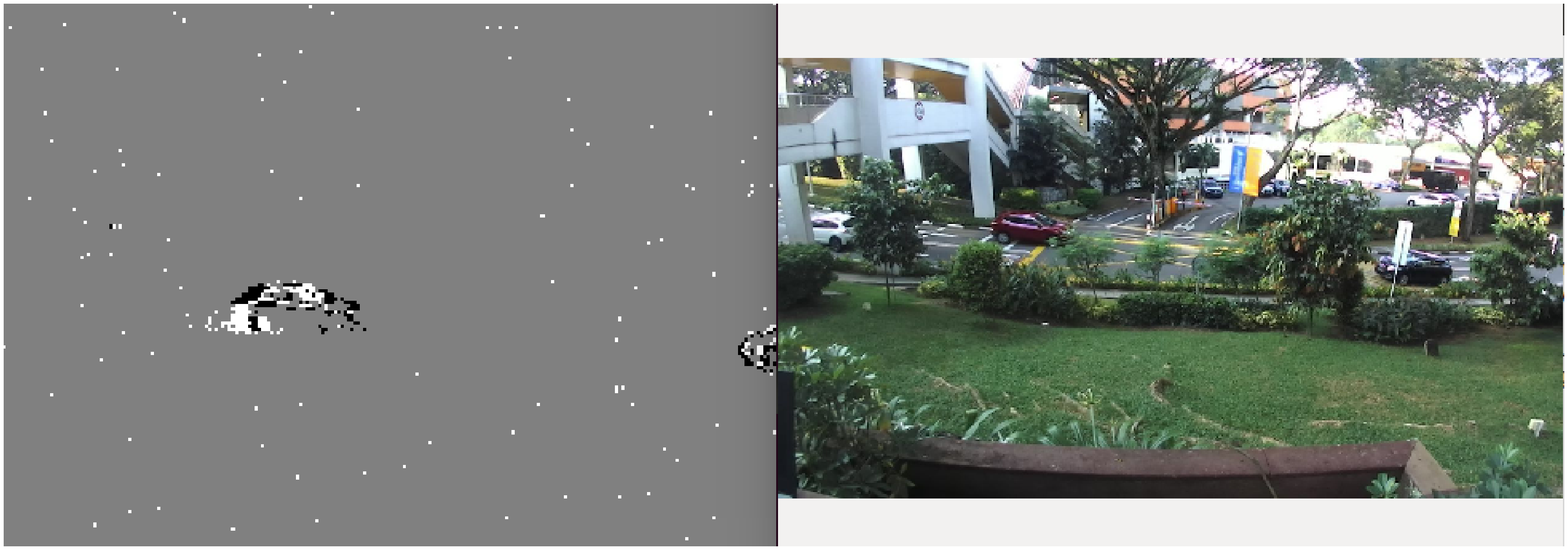}}}
\caption{Examples of recorded event-based data.}
\label{fig:events_eg}
\end{figure}

The development of this work demanded the acquisition of event-based data from a real application scenario for the purposes of training, validation and testing of our system. The main requirement for these recordings was a high, perpendicular view from the road near intersections. Under this condition, three places inside our campus were chosen for data recording (samples shown in Fig.~\ref{fig:events_eg}). The events are aggregated to generate a frame either every 66 ms. During trial-and-error experiments, surprisingly even longer time periods (100ms) did not degrade system performance, although it increases the latency of the system significantly. The setting of 66 ms for generating the frame was chosen as a trade-off between accuracy and latency. 

\par
Table~\ref{tab:data_overview} shows the distribution of the collected dataset in terms of the number of samples obtained for each category. We noticed that there were a lot of car samples, and thus to balance the training data, the samples were augmented by random flipping, rescaling (up to 140\%) and rotation (up to 20 degrees in either direction). After augmenting a sample, it was cropped back to 42\x42 pixels for training with a fixed image size on TrueNorth. A separate test dataset captured at a different time was used for evaluating the system. 
\par
To provide critical insights on how a standard frame-based setup may perform for the same application, RGB data was recorded using a ZED camera simultaneously along with the events captured by the DAVIS camera. For both the RGB and events datasets, manual annotation was carried out to facilitate tracker and classifier evaluation. 
\begin{table}[t]
\caption{Number of samples per category in the collected dataset.}
\centering
\begin{tabular}{rrrrrr}
\hline\hline
~ &      Car & Bus   & Pedestrian & Bike & Truck/Van  \\[0.5ex]
\hline
Site 1 & 322     & 30    & 115        & 43   & 18       \\
Site 2 & 226     & 105   & 53         & 14   & 28       \\
Site 3 & 390     & 181   & 89         & 39   & 56       \\
Sum & 938     & 316   & 257        & 96   & 102   \\
\%  & 54.89   & 18.49 & 15.04      & 5.62 & 5.97        \\[1ex]
\hline
\end{tabular}
\label{tab:data_overview}
\end{table}

\subsection{Comparison to State-of-the-art} 
\label{sec:statecompare}

In this section, we first report the performance of the proposed EOT, Continuous-EOT trackers and compare it to the popular multi-object event-based mean shift (EBMS) tracker and conventional Kalman filter (KF) tracker. Additionally, we also evaluate our tracker performance on recordings from different neuromorphic vision sensors. Next, we compare the classification performance of the TrueNorth model against state-of-the-art method DART \cite{Ramesh2019} and SLAYER \cite{shrestha2018slayer} to investigate whether there is a performance drop due to the binary frame generation process in our neuromorphic framework. For a direct comparison between events and RGB data, we report the tracking and classification performance on simultaneously recorded RGB data compared to events. Finally, we show how the TN EEDN hyperparameter constraints affect the classification performance compared to fully-trained CNNs on the ImageNet database via transfer learning.

\subsubsection{Comparison of Tracker Performance}
\label{sec:trackperf}
\begin{figure}
\centering     
\subfloat[]{\includegraphics[width=85mm]{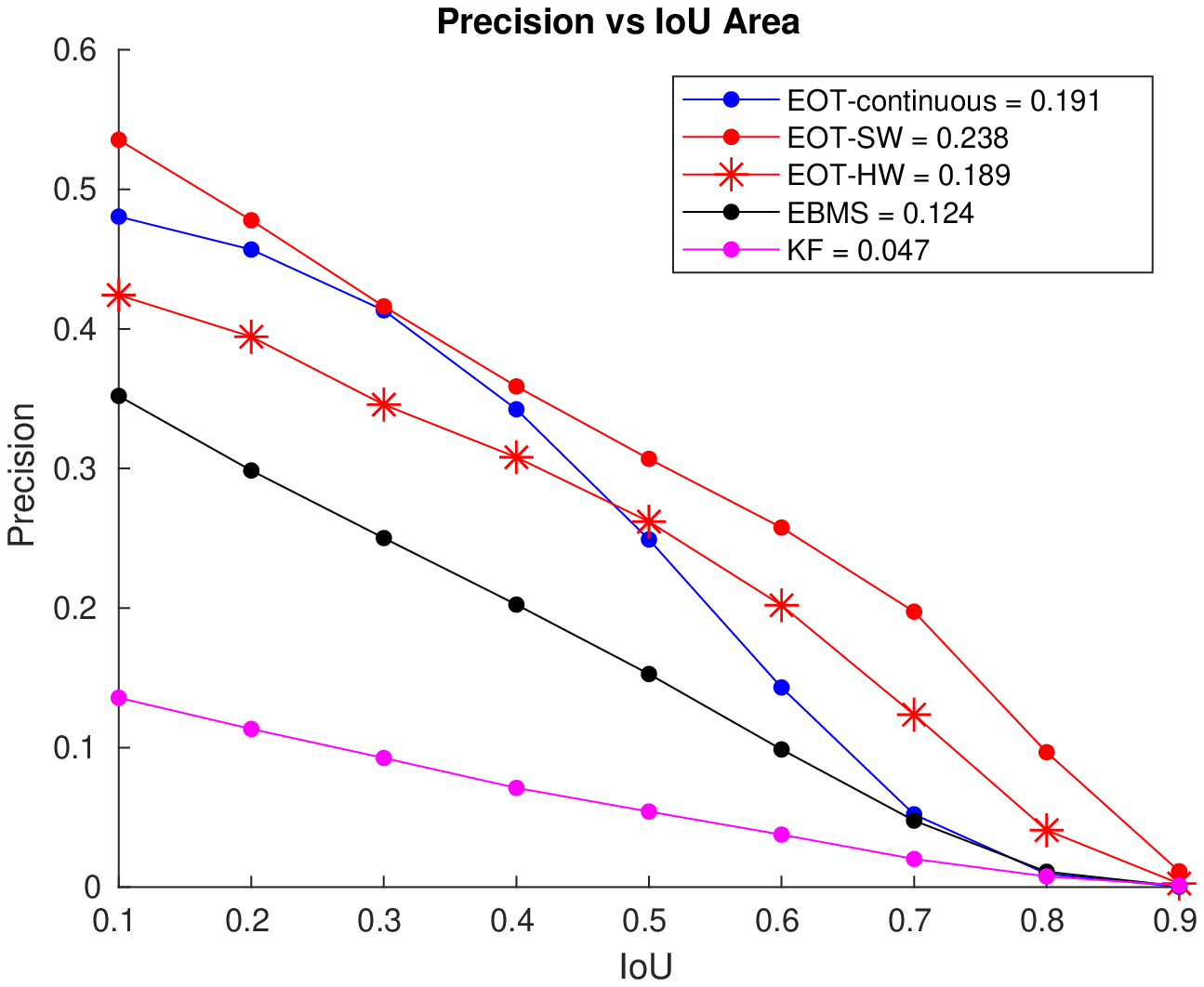}}
\vspace{0.1em}
\subfloat[]{\includegraphics[width=85mm]{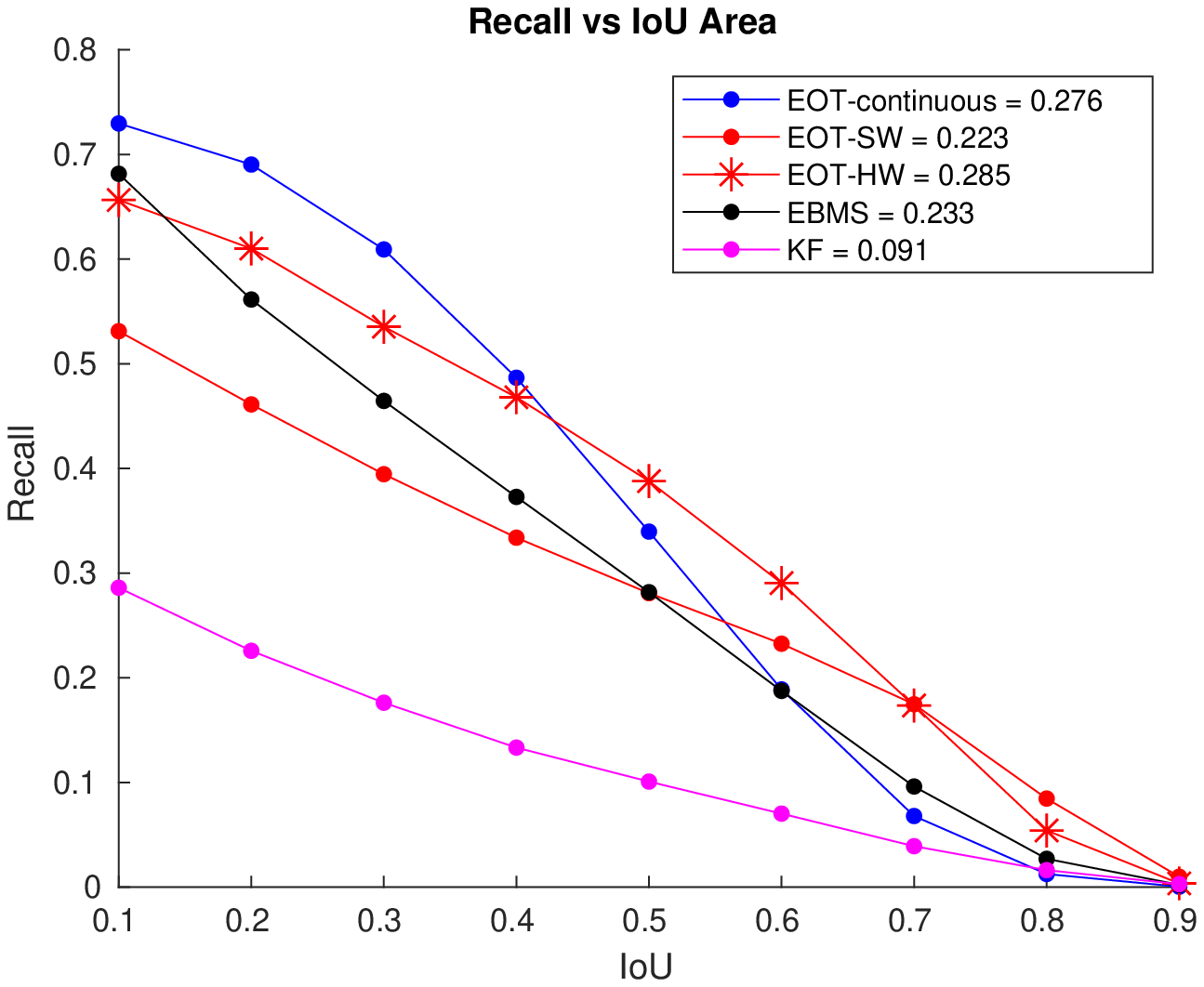}}
 \caption{Comparison of the proposed EOT tracker -- in software (SW) and hardware (HW), and Continuous-EOT tracker -- to Kalman Filter (KF) tracker on event data \cite{Lin2015} and Event-Based Mean-Shift (EBMS) \cite{Delbruck2013} in terms of precision and recall at different IoU thresholds.}
\label{fig4:all_cases_PR}
\end{figure}

To analyze the effect of finite bit precision in hardware implementations, we extracted the region proposals from the FPGA and passed it through the software tracker to generate precision and recall curves at the different intersection over union (IoU) thresholds, following the protocols in \cite{Authors}. 
Fig.~\ref{fig4:all_cases_PR} compares the proposed EOT and the continuous-time EOT tracker performance to the multi-object event-based mean shift (EBMS) tracker \cite{Delbruck2013} and a Kalman Filter (KF) tracker used in \cite{Lin2015}. For initializing the Kalman Filter tracker, an initial bounding-box location using the manual track annotations is given as input for each object. Note that in the Continuous-EOT tracker, there is no frame creation as the input to the tracker and is computationally faster. Fig.~\ref{fig4:all_cases_PR} also shows the performance of the proposed EOT tracker in software and hardware. 
\par
It can be seen that the EOT tracker outperforms the multi-object EBMS tracker \cite{Delbruck2013} comfortably (7\x less memory and 3\x less computations \cite{Authors}). In terms of overall F1-score combining the precision and recall statistics, the EOT-software tracker scores 0.35 compared to 0.21 for the EBMS. Additionally, the EOT tracker performs better than the KF tracker applied to event-based binary frames. Note that EOT hardware has a slightly lower F1-score of 0.3 due to finite precision arithmetic. It can be deduced that the Continuous-EOT tracker, which processes each event individually, has similar performance to EOT in low IOUs and performs gradually worse as IOU increases. Since the latency is much lower, it can be potentially used as a real-time intrusion detection system for fast moving objects, as the frame-generation process in EOT is avoided. 

\par
\subsubsection{Comparison to other NVS}
The proposed EOT method was also evaluated on recordings from different neuromorphic vision sensors. The following event-based cameras were used for this evaluation: a CeleX-V from CelePixel, a DAVIS 640 and a DAVIS 240C, both from Inivation. Their resolution is $1280\times800$, $640\times480$ and $240\times180$, respectively. 

Recordings of ongoing traffic were performed with all three sensors simultaneously for one hour. Subsequently, each recording was manually annotated to create ground truth tracks for object tracking evaluation. This data was evaluated in terms of precision and recall metrics, as introduced in Section \ref{sec:trackperf}, and in regards to detection probability. In the latter, an object is considered successfully detected if the IoU of the tracker box is greater than the threshold in at least one point in time throughout its path, i.e. if the tracker generates at least one true positive box for that object.
\par
\begin{figure}
\centering     
\subfloat[]{\includegraphics[width=85mm]{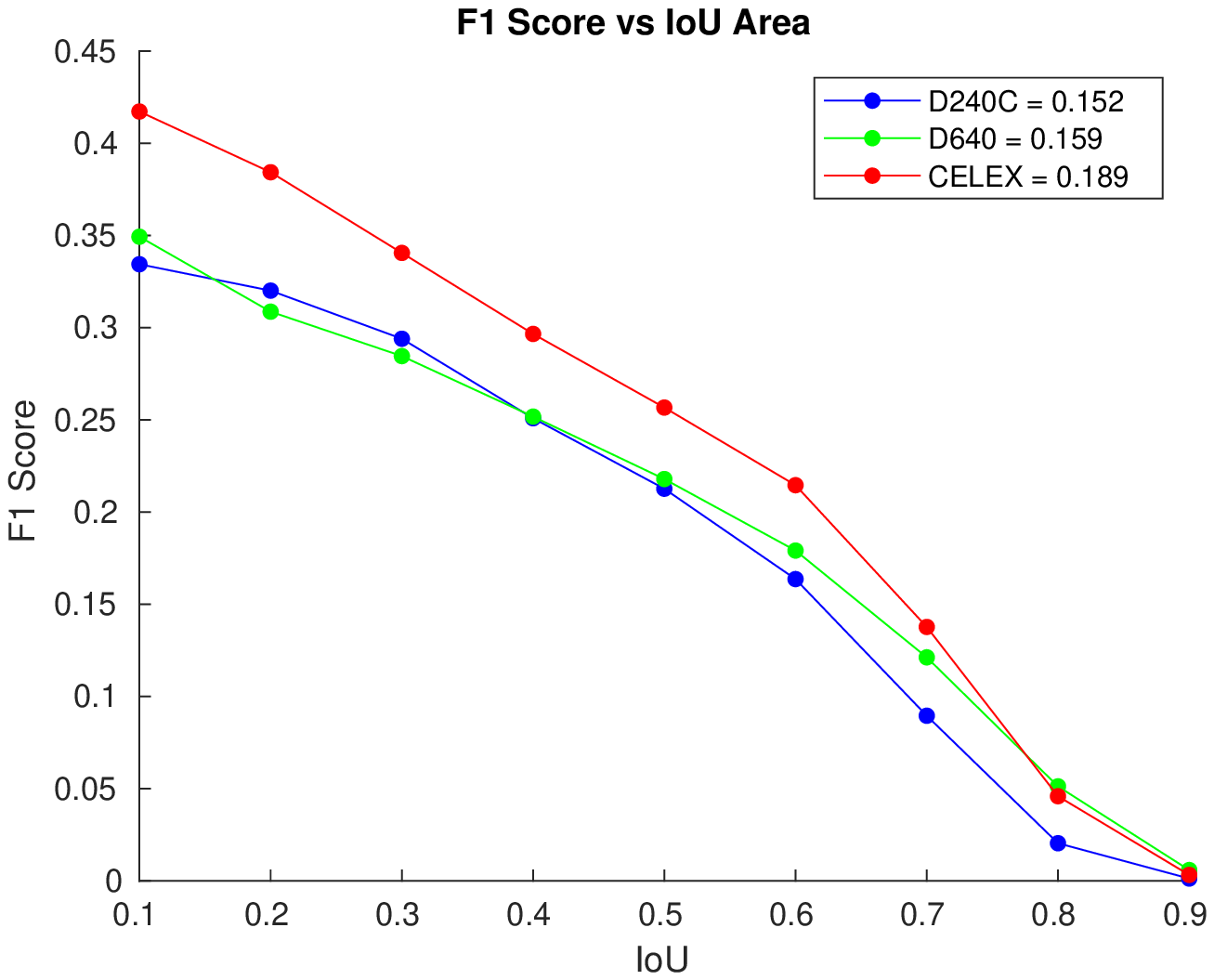}}
\vspace{0.1em}
\subfloat[]{\includegraphics[width=85mm]{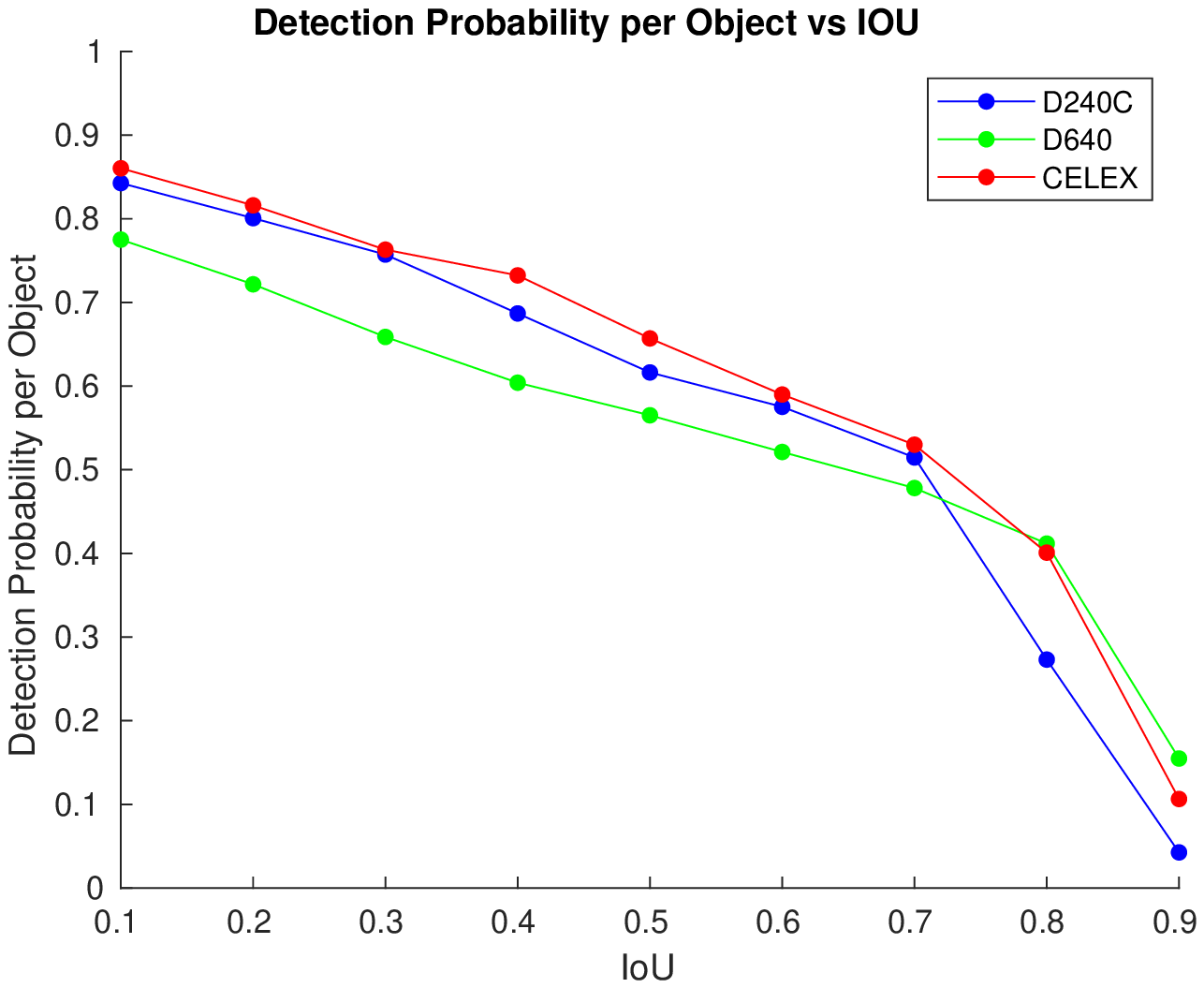}}
 \caption{Comparison of EOT using recordings from different NVS (DAVIS 240C, DAVIS 640 and CELEX) in terms of F1 Score for different IoU thresholds.}
\label{fig8:NVS}
\end{figure}


\begin{table*}[t]
\caption{Comparison of TrueNorth (TN) classification performance to  state-of-the-art methods.}
\centering
\begin{tabular}{|l|r|r|r|r|r|r|}
\hline
Method &      Car\% & Bus\%   & Human\% & Bike\% & Truck\% & Balanced\%\\[0.5ex]
\hline\hline
TN(Per-Sample)  & 90.4 & 92.5  & 94.7  & 86.9 & 54.2  & 83.3       \\
TN(Per-Track)  & 99.0   & 98.2  & 100   & 100  & 53.8  & 90.2       \\
TN RGB(Per-Track)  & 99.4   & 98.0  & 100   & 95.5  & 72.5  & 93.1  \\
DART (Per-Track)  \cite{Ramesh2019} & 96.3     & 96.6   & 100         & 100   & 83.9    & 95.4   \\
SLAYER (Per-Track)  \cite{shrestha2018slayer} & 97.6     & 98.3   & 100         & 100   & 78.1    & 94.8   \\
RESNET-18  \cite{DBLP:journals/corr/HeZRS15}  & 94.8   & 93.1  & 99.0   & 100  & 87.0  & 94.8       \\
RESNET-18 (RGB)  \cite{DBLP:journals/corr/HeZRS15}  & 99.4   & 98.0  & 100   & 100  & 82.5  & 96.0       \\[1ex]
\hline
\end{tabular}
\label{tab:test_accuracy_GT}
\end{table*}

Performance results in terms of the F1 score are shown in Fig. \ref{fig8:NVS}a. It can be observed that all sensors follow a similar trend with respect to the IoU axis. Overall, the Celex sensor performs better, showing good results in both low and high IoU values. Additionally, a correlation can be seen between the sensors' spatial resolution and tracking efficacy on IoU values greater than $0.6$. This can be attributed to the better ability of sensors with the larger resolution to capture more data, allowing the Celex camera to collect objects moving at low speed like humans.

Results for detection probability are shown in Fig. \ref{fig8:NVS}b. All sensors present positive results for this metric, showing a probability above $60\%$ for IoU values under $0.5$. Further, the Celex sensor performs marginally better for most of the IoU values, confirming the intuition that a higher resolution sensor will have a higher probability of detecting an object in its field of view.

\subsubsection{Comparison to state-of-the-art classification methods}
Table~\ref{tab:test_accuracy_GT} shows the TrueNorth test accuracy evaluated using the ground truth (not using the tracker output) under two settings:  per-sample (each instance of an object) and per-track (majority voting across all instances of the same object). As expected, there is a significant improvement in the accuracy when considered on a per-track basis. Nonetheless, the TrueNorth model struggles with Trucks due to their similarity to both buses and cars, but does very well on all other classes, especially when given multiple opportunities to classify them on a per-track basis. 
\par
However, it is imperative that for a real-world surveillance application, the back-end classifier is trained on representative samples from the tracker output rather than on manually annotated ground truth tracks. The tracker output does not have object labels, and in order to train the TrueNorth classifier, class labels are automatically generated using their overlap with the ground truth object locations. The spurious tracks with no ground truth overlaps can be labelled as an additional background class and subsequently TrueNorth can classify them as false positives during deployment. Table~\ref{tab:test_accuracy} shows the TrueNorth classification accuracies obtained on the test track output (not the ground truth test tracks) by three models trained on: (1) ground truth track outputs, (2) augmented ground truth track outputs, and (3) the hardware tracker outputs using the training data. We see that the models trained on ground truth tracks that contain no background class perform poorly. As mentioned earlier in the introduction, this is an important highlight of our proposed system, to be able to respond to moving background conditions or spurious tracker outputs. In particular, the model trained on the tracker output gets a much higher accuracy (70.4\%) compared to models trained on ground truth tracks.
\par
In practice, the above-mentioned system accuracy will be higher due to limitations in evaluating the tracker outputs obtained by auto-generated class labels from the manually annotated ground truth. Trackers that do not have overlap with their target can only be labelled as background and this is especially true for the case of an object leaving/entering the scene when manual annotations do not exist. Thus, when the tracker locks on before the ground truth tracks have started, TrueNorth correctly classifies it as a bus, but since it does not agree with the annotation, it is marked as an error in the evaluation (Table \ref{tab:test_accuracy}). Similarly, when the object exits the scene, TN correctly classifies it as a bus but it is counted as a type II error because of the mismatch with the ground truth annotation. A demo of our fully embedded system reveals this scenario clearly\footnote{Video demo (updated): \url{https://tinyurl.com/ycc2tn5t}}.

\begin{table}
\centering
\caption{Type of Training Data vs. Per-Track TrueNorth Test Accuracy.}\label{tab:test_accuracy}
\begin{tabular}{|l|c|r|r|r|}
  \hline
  Class & Name  & GT \% &  Aug. GT \% & FPGA \%\\ \hline
  1 & Human     & 91.8   & 96.3 & 85.8\\
  2 & Bike      & 62.2   & 72.9 & 50.0\\
  3 & Car       & 62.0   & 56.7 & 89.8\\
  4 & Truck     & 22.7   & 35.1 & 28.5\\
  5 & Bus       & 53.1   & 54.0 & 84.2\\
  6 & Other     & 0      & 0    & 83.9\\
    & Overall   & 39.7   & 39.1 & 82.6\\
    & Balanced  & 48.7   & 52.5 & 70.4\\
  \hline
\end{tabular}
\end{table}


Table~\ref{tab:test_accuracy_GT} compares the TrueNorth classification accuracy to the  Distribution Aware Retinal Transform (DART) framework \cite{Ramesh2019}, which has obtained state-of-the-art accuracy on multiple event-based object datasets. It is worth stating that the DART method utilizes all the event information and obtains 95.4\% per-track accuracy on the ground truth test dataset. This is understandably higher than the per-track 90.2\% accuracy obtained using the EEDN framework, but more importantly, our approach shows that event cameras can be utilized to generate ``frames'' without sacrificing much performance for embedded surveillance applications. 
\par

Table~\ref{tab:test_accuracy_GT} also compares the classification performance of TrueNorth and state-of-the-art learning algorithm Spike layer error reassignment in time (SLAYER) \cite{shrestha2018slayer}, which learns both weight and axonal delay parameters in Spiking Neural Networks. The classification performance of SLAYER shows a similar trend as the DART method, higher than TrueNorth classification performance. It is interesting to observe that the difference in balanced accuracy mainly depends on the classifier performance on Trucks, whereas the difference in accuracy's in other classes across various methods is negligible. In other words, fine-grained classification needs to exploit subtle inter-class object appearance variations, which can be captured by event-based methods exploiting high temporal resolution of the NVS. 


\subsubsection{Comparison with RGB} 
\begin{figure}[t]
\centering     
%
%
\includegraphics[width=85mm]{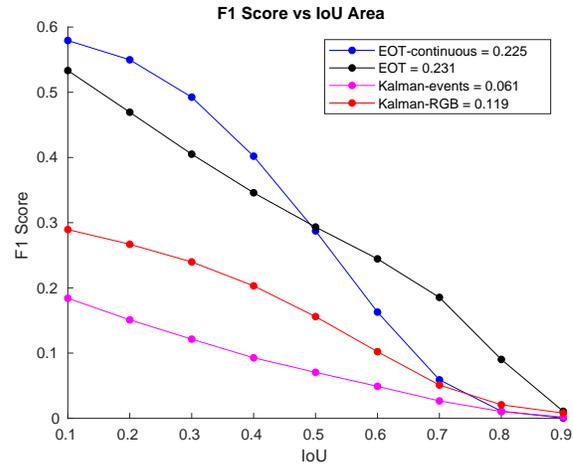}
\vspace{0.1em}
 \caption{Comparison of EOT and Continuous-EOT to Kalman Filter (KF) tracker on RGB data in terms of F1 Score for different IoU thresholds.}
\label{fig6:all_cases_RGB_PR}
\end{figure}

For a close comparison of the proposed EOT and Continuous-EOT trackers to the KF tracker \cite{Lin2015}, the method was tested on standard RGB data recorded as they were not developed for event-based binary frames. Fig.~\ref{fig6:all_cases_RGB_PR} compares the tracking performances on RGB data. Both Continuous-EOT and EOT tracker again outperforms the frame-based tracker with a higher F1-score of 0.35. 
 However, the EOT tracker's overall performance is slightly better, with an F-score of 0.231 compared to 0.225 for Continuous-EOT. 
As expected, the Kalman Filter implementation performs better on RGB data when compared to event-based data. The KF-based multi-object tracker scores 0.119 on RGB and only 0.061 for events data.
\par
\begin{figure}[t]
\centering     
\includegraphics[width=80mm]{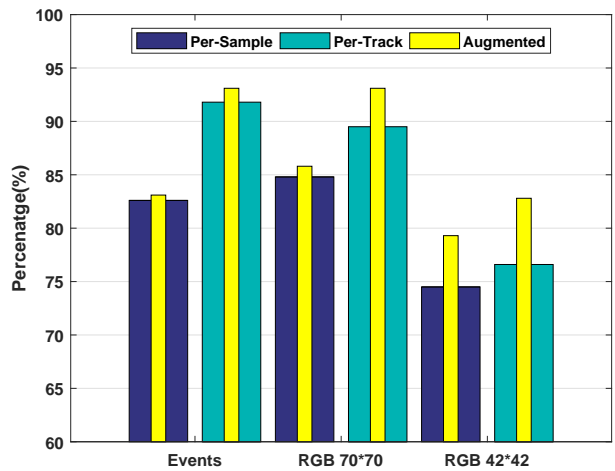}
\vspace{0.1em}
 \caption{{Comparison of TrueNorth classification performance on events and RGB datasets.}}
\label{fig6RGB}
\end{figure}

To prove the classification performance on the dataset recorded by the event camera is in the same ballpark with dataset simultaneously recorded by standard RGB Camera, we compare the TrueNorth classification model performance on the events and RGB datasets. Note that the RGB dataset has the same train and test split as the events data. Fig.~\ref{fig6RGB} shows the TN performance of Events and RGB 70$\times$70 are in the same neighbourhood whereas RGB 42$\times$42 shows a significant performance drop. This also implies that higher image resolution better the classification performance on TrueNorth. Augmented datasets show a slight increase in performance compared to unbalanced ones. In particular, the CNN inference on TrueNorth for a 70$\times$70 RGB image utilizes 3721 out of the available 4096 cores, which is 1.4 times the number of cores used by the events data.
%

\subsubsection{Comparison to Pre-trained Models} 
\begin{figure}
\centering     
\includegraphics[width=80mm]{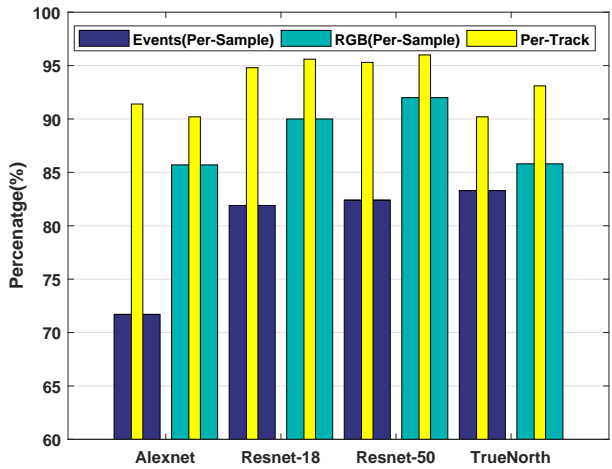}
\vspace{0.1em}
 \caption{{{Comparison of classification performance of events and RGB data-sets via Transfer Learning}}}
\label{fig7}
\end{figure}

RGB and Events ground truth data are streamed into pre-trained networks such as Alexnet \cite{krizhevsky2012imagenet}, Resnet-18 and Resnet-50 \cite{DBLP:journals/corr/HeZRS15} via transfer learning to see how TrueNorth performs compared to CNNs that have more number of layers and having been trained on over a million images. Recall that TrueNorth uses a 15-layer CNN with hardware constraints on the hyper-parameters \cite{Esser2015}.  Figure.~\ref{fig7} shows a gradual increase in accuracy from Alexnet to Resnet-18  and Resnet-50 for both RGB as well as Events datasets. RGB performs slightly better than events, which can be attributed to having more visual information such as color, texture, etc. Nonetheless, TN achieves a comparable performance to pre-trained CNNs with deeper architectures on both events and RGB datasets. 

%% file: 05_conclusion.tex
\section{Conclusion}
This paper presented one of the first end-to-end neuromorphic frameworks for real-time object tracking and classification demonstrated using a low-power hardware implementation that consumes about 0.5W, which is 3\x less power than conventional TPUs used for deep learning and 4\x lesser than state-of-the-art frame-based embedded systems for real-time tracking. The proposed framework employs a hybrid approach consisting of events aggregated into frames for maintaining individual track of objects in occluded scenarios. Subsequently, the tracked object was efficiently classified using the IBM EEDN pipeline of the spike-based neuromorphic chip. In this setup, the TrueNorth chip was time-multiplexed to handle eight objects while making sure that the neurons used for pre-processing will scale linearly with the number of objects to be pseudo-simultaneously classified. As an optional continuous-time implementation, we demonstrated the use case of the proposed event-based tracker to fully exploit the low latency characteristics of the NVS. In addition to a real-time demo, we extensively compared the proposed tracking and classification methods to state-of-the-art event-based and frame-based methods and showed its relevance to on-going work in the research field. We also demonstrated that the proposed neuromorphic system achieves better tracking and classification performance compared to a standard RGB camera setup when simultaneously evaluated over several hours of traffic recordings. In summary, we have demonstrated a strong use case of our neuromorphic framework for real-time and embedded applications without sacrificing performance.

%% file: 06_biography.tex
\begin{IEEEbiographynophoto}
{Andres Ussa}
received his B.Sc. degree in Mechatronics Engineering from Nueva Granada Military University in 2012 and Joint M.Sc. in Embedded Computing Systems from TU Kaiserslautern and University of Southampton in 2016. His previous research experience has been focused on embedded systems design and machine learning applications. He had a short experience as a Software/Hardware Developer for consumer electronics. 
\end{IEEEbiographynophoto}
\vspace{-1 cm}
\begin{IEEEbiographynophoto}
{Chockalingam Senthil Rajen}
is currently pursing his final year  B. Tech in National Institute of Technology Tiruchirappalli, India . His research work has been focused on Computer vision and Deep Learning applications. He has a notable research experience in the field of robotics involving motion planning and multi agent systems.
\end{IEEEbiographynophoto}
\vspace{-1 cm}
\begin{IEEEbiographynophoto}
{Deepak Singla}
(M’ 20) received his B.Tech in Electrical Engineering (Power \& Automation) from the Indian Institute of Technology, Delhi in 2018. After the graduation, he joined Nanyang Technological University, Singapore as a Project Officer in the BRAIN Systems Lab - CICS headed by Dr. Basu. Since then, he is working on IoT based applications of neuromorphic vision sensors and designing low computational cost and reliable systems for the same. His research interests include neurmorphic engineering, brain-machine intelligence and computer vision. 
\end{IEEEbiographynophoto}

\vspace{-1 cm}
\begin{IEEEbiographynophoto}
{Jyotibdha Acharya}
received his B.E. degree in ECE from Jadavpur University, India in 2014. He is currently working toward the Ph.D. degree at HealthTech NTU, Interdisciplinary Graduate Program, Nanyang Technological University, Singapore. Previously, he worked in Philips Healthcare as a Graduate Engineer Trainee (Imaging Systems). His research interests include neuromorphic alogorithms, deep learning, and biomedical applications.
\end{IEEEbiographynophoto}
\vspace{-1 cm}
\begin{IEEEbiographynophoto}
{Gideon Fu Chuanrong}
is a senior undergraduate student of the School of Computing Department at the National University of Singapore. His research focuses on Computer Vision, Image Processing, and Signal Processing. He had the experience of research internship at the N.1 Institute for Health, NUS, Singapore as part of this work.
\end{IEEEbiographynophoto}
\vspace{-1 cm}
\begin{IEEEbiographynophoto}
{Arindam Basu}
(M’ 10, SM' 17) received the B.Tech and M.Tech degrees in Electronics and Electrical Communication Engineering from the Indian Institute of Technology, Kharagpur in 2005, the M.S. degree in Mathematics and PhD. degree in Electrical Engineering from the Georgia Institute of Technology, Atlanta in 2009 and 2010 respectively. Dr. Basu received the Prime Minister of India Gold Medal in 2005 from I.I.T Kharagpur. He joined Nanyang Technological University in June 2010 and currently holds a tenured Associate Professor position. 

He is currently an Associate Editor of IEEE Sensors journal, IEEE Transactions on Biomedical Circuits and Systems and Frontiers in Neuroscience. He was a Distinguished Lecturer for IEEE Circuits and Systems Society for the 2016-17 term. Dr. Basu received the best student paper award at Ultrasonics symposium, 2006, best live demonstration at ISCAS 2010 and a finalist position in the best student paper contest at ISCAS 2008. He was awarded MIT Technology Review's inaugural TR35@Singapore award in 2012 for being among the top 12 innovators under the age of 35 in SE Asia, Australia and New Zealand. His research interests include bio-inspired neuromorphic circuits, non-linear dynamics in neural systems, low power analog IC design and programmable circuits and devices.
\end{IEEEbiographynophoto}
\vspace{-1 cm}
\begin{IEEEbiographynophoto}
{Bharath Ramesh}
 works mainly in the research areas of pattern recognition and computer vision. At present, his research is centered on event-based cameras for autonomous sensing and navigation. This includes tracking and recognition with an array of sensors, most importantly event-based cameras, to be processed efficiently on low-power devices to yield accurate results in real-time. In the past, he worked on object recognition and related areas such as scene understanding, face recognition, and object detection for silicon retinal, event-based cameras on-board unmanned aerial vehicles. He received the B.E. degree in electrical \& electronics engineering from Anna University of India in 2009; M.Sc. and Ph.D. degrees in electrical engineering from National University of Singapore in 2011 and 2015 respectively, working at the Control and Simulation Laboratory on Image Classification using Invariant Features. 
\end{IEEEbiographynophoto}
\vfill